%% file: main.tex
\documentclass[letterpaper]{article}
\usepackage[preprint]{aaai2027}
\usepackage[hyphens]{url}
\usepackage{graphicx}
\usepackage{natbib}
\usepackage{caption}
\DeclareCaptionStyle{ruled}{labelfont=normalfont,labelsep=colon,strut=off}

\usepackage{booktabs}
\usepackage{amsmath}

\usepackage[colorlinks=true,linkcolor=black,citecolor=black,filecolor=black,urlcolor=blue]{hyperref}
\makeatletter
\expandafter\let\csname ver@hyperref.sty\endcsname\relax
\makeatother

\title{Tunable Tool-Call Rates in LLM Agents via Representation Steering}
\author{
    Yuqi Chen\textsuperscript{\rm 1},
    Vincent Siu\textsuperscript{\rm 1},
    Yang Liu\textsuperscript{\rm 1},
    Dawn Song\textsuperscript{\rm 2},
    Chenguang Wang\textsuperscript{\rm 1}\corresponding
}
\affiliations{
    \textsuperscript{\rm 1}Department of Computer Science and Engineering, UC Santa Cruz\\
    \textsuperscript{\rm 2}Department of Computer Science, UC Berkeley\\
    \{yche1052, vsiu3, yangliu, chenguangwang\}@ucsc.edu \quad dawnsong@cs.berkeley.edu
}

\newcommand{\vince}[1]{}
\newcommand{\yuqi}[1]{}

\begin{document}

\maketitle


\input{text/abstract}

\input{text/intro}

\input{text/method}

\input{text/experiment}

\input{text/vecanalysis}
\input{text/rw}
\input{text/conclusion}

\bibliography{aaai2027}

\clearpage
\appendix
\input{appendix}

\end{document}

%% file: text/abstract.tex
\begin{abstract}
Deciding whether to call a tool is a core competence of an LLM agent, and a costly one to get wrong: needless calls add latency, accrue cost, and may trigger irreversible side effects, while missing calls leave the model confidently wrong on questions it could only answer through tool-calls. Models manage this balance poorly, both over-using and under-using tools. Existing methods such as post-training and prompt engineering are expensive and difficult to modify at inference time.  We show that whether an instruction-tuned model calls a tool can be controlled by a single linear direction in its residual stream, extracted without any training from the model's own tool-use preference signal and turned into an inference-time intervention with no prompt change. Adding the direction with strength $\alpha$ moves the call rate monotonically from near $0\% $ to over $90\%$ while keeping calls well-formed. The steering works in both directions: dialing it down suppresses calls, and dialing it up induces new calls that land precisely on the questions the model cannot answer from its own knowledge. We also show that the direction generalizes to unseen tools with strength comparable to each tool's own direction and without favoring any specific tool choice. With live tool execution, a single sweep of the steering traces a cost/accuracy Pareto frontier and nearly doubles open-domain QA accuracy ($0.29 \! \rightarrow \! 0.56$); the same recipe transfers across a diverse range of models spanning dense, MoE, and multimodal architectures, without any training.  Our code is publicly available at \url{https://github.com/YuqiChen4188/Steering-Tool-Use-Propensity}.


\end{abstract}

%% file: text/intro.tex
\section{Introduction}

\begin{figure}[t]
\centering
\includegraphics[width=0.48\textwidth]{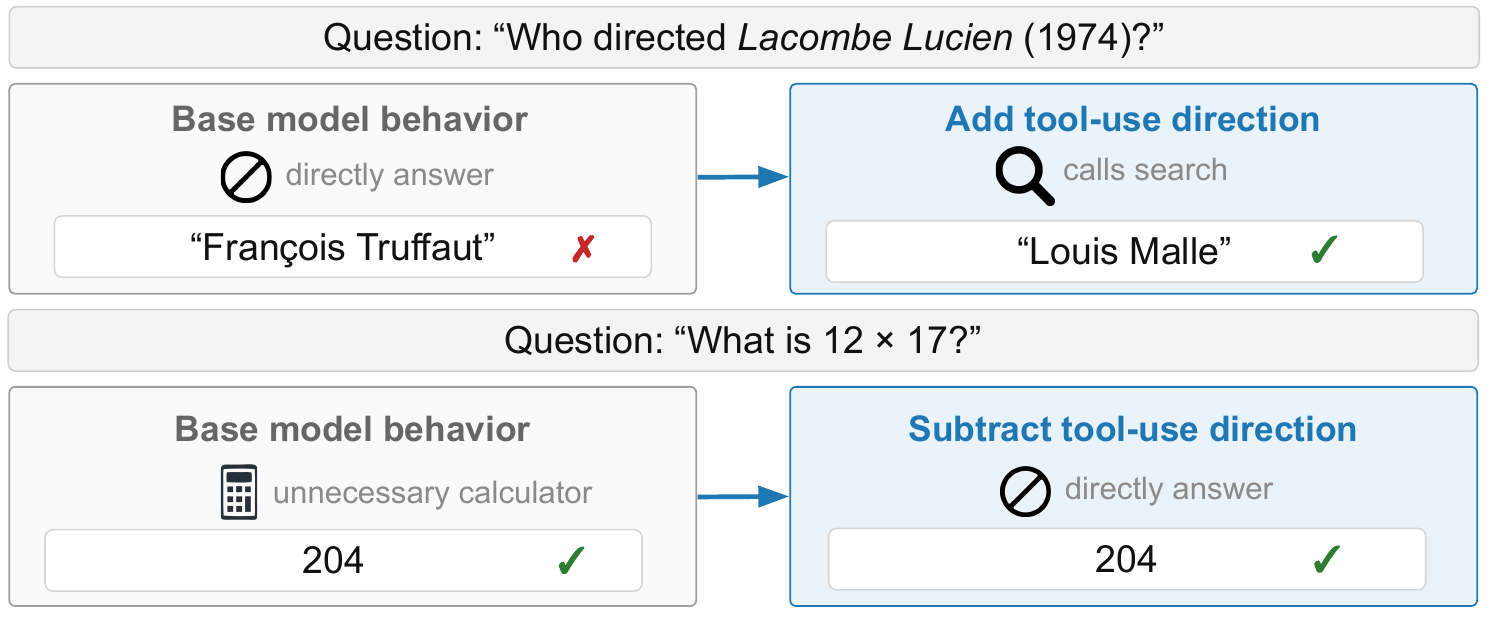}
\caption{Bidirectional control of tool use through representation steering. For an uncommon factual question, the unsteered model answers incorrectly without using a tool. Adding the tool-use direction induces a search call and leads to the correct answer. For a math question that the base model can already solve, it calls the calculator unnecessarily, while subtracting the direction removes the needless call and the answer stays correct.}
\label{fig:teaser}
\end{figure}

When an LLM agent chooses to use a tool~\citep{schick2023toolformer, mialon2023augmentedlanguagemodelssurvey}, and how readily it does so, directly affects its cost and reliability. 
Each call spends latency and money and, for action tools such as sending an email or executing code, can cause irreversible side effects. Each skipped call risks a confidently wrong answer to a question the model could only get right by looking it up. As industry moves toward large-scale agentic deployments, exemplified by NVIDIA and OpenAI's multi-gigawatt infrastructure partnership~\citep{nvidiaopenai2025partnership}, these costs are borne at deployment scale rather than in the lab. 
However, agents manage this balance poorly in both directions. As shown in Fig.~\ref{fig:teaser}, they \emph{under-use} tools, answering long-tail factual questions incorrectly from memory rather than calling a tool~\citep{mallen-etal-2023-trust}. They also \emph{over-use} tools that add cost without influencing correctness~\citep{huang2024metatoolbenchmarklargelanguage, ning2024wtuevalwhetherornottoolusage}. This motivates a lightweight way to control tool-calling behavior at inference time without retraining the model or rewriting the prompt.

Prior work on tool-use mitigation is expensive and hard to customize at inference time. Existing works make agents more selective about when to use tools through fine-tuning \citep{qian-etal-2025-smart, wang-etal-2025-self, shen-etal-2024-smartcal}, and retrieval-augmented systems~\citep{lewis2020retrieval} learn when to retrieve from supervision \citep{mallen-etal-2023-trust}. Meanwhile, interpretability methods have begun to open up tool use from the inside. For instance, the choice of tool can be linearly represented~\citep{park2024linearrepresentationhypothesisgeometry} and can be steered by adding a direction to the residual stream \citep{wu2026toolcallinglinearlyreadable}. However, these results leave open whether the preceding binary decision to use a tool at all has a similarly readable and steerable internal representation.
Such a representation could provide direct inference-time control over an agent's tool reliance without retraining or modifying its prompt.

To address this question, we show that the tool-calling decision can be controlled by a single direction in representation space across different models. We identify this direction via difference-of-means (DIM), and demonstrate bidirectional control over a model's tendency to call tools by both encouraging and limiting tool use. The resulting steering is knowledge-selective, transfers to tools absent from the extraction harness, and changes whether a tool is called without substantially affecting which tool is selected. With live tool execution, varying the intervention strength traces a cost--accuracy Pareto frontier and raises open-domain QA accuracy from $0.29$ to $0.56$ at roughly one search call per question.

We summarize our main contributions as follows:
\begin{itemize}
\item We identify a single linear direction in the residual stream that steers whether a model calls a tool, providing continuous inference-time control without training or prompt modification. 
\item We show that this control is knowledge-selective and tool-general. A direction extracted in the multi-tool harness steers held-out tools at or near the strength of their individually extracted directions, while remaining distinct from the direction that controls which tool is selected in a multi-tool setting.
\item We demonstrate the practical utility and generality of this control. With live tool execution, varying its strength traces a cost--accuracy Pareto frontier and nearly doubles open-domain QA accuracy. The same procedure generalizes across a diverse range of models.
\end{itemize}

%% file: text/method.tex
\section{Method}
\label{sec:method}

Our pipeline has three training-free ingredients: a cheap forward-pass measure of tool-use propensity, a difference-of-means direction extracted from that measure in a multi-tool environment, and an inference-time steering intervention that adds the direction back at inference. Throughout, we study tool use in a multi-tool harness: a chat template presents the model with a list of available tools through its native tool-calling format, under a neutral system prompt that permits but does not require tool use. We extract the steering direction in this multi-tool setting, never in a single-tool prompt, so that the contrastive split reflects the decision to call any tool rather than a feature of one specific tool. The tools, datasets, and models that instantiate this environment are described in \S\ref{sec:setup}.

\subsection{A forward-pass propensity proxy}
\label{sec:proxy}
We read the tool-use preference signal from the model's own tool-calling format. The assistant's call opens with a single, tool-exclusive special token $t^{\star}$ (\texttt{<tool\_call>}) shared by every available tool, so that $P(t^{\star})$ is a tokenization-clean measure of whether the model will call anything. For a query $q$ rendered in the multi-tool harness as $x(q)$, we score propensity by the log-probability that the model emits $t^{\star}$ at the first generated position,
\begin{equation}
s(q) \;=\; \log p_\theta\!\big(t^{\star} \mid x(q)\big),
\label{eq:proxy}
\end{equation}
which requires a single forward pass and no generation. On held-out queries, $s(q)$ is a faithful ranking signal for real tool use; because some calls occur mid-generation, we use real generation whenever absolute call rates matter. Equation~\eqref{eq:proxy} turns the layer/magnitude search below into a batch of forward passes.

\subsection{Extracting the direction}
\label{sec:extract}
We rank the queries in the multi-tool harness by $s(q)$ and form a high-propensity set $\mathcal{H}$ (top quantile) and a low-propensity set $\mathcal{L}$ (bottom quantile). At each layer $\ell$, writing $h_\ell(q)$ for the residual stream at the last prompt token, the steering vector is the difference of pool means,
\begin{equation}
v_\ell \;=\; \frac{1}{|\mathcal{H}|}\sum_{q\in\mathcal{H}} h_\ell(q) \;-\; \frac{1}{|\mathcal{L}|}\sum_{q\in\mathcal{L}} h_\ell(q).
\label{eq:dom}
\end{equation}
Because $t^{\star}$ opens a call to any available tool, the high-propensity pool is a near-even mix of the different question types, so $v_\ell$ captures a general call-propensity rather than a tool-specific one. No labels, gradients, or sparse autoencoders are used; a few thousand queries suffice.

\subsection{Inference-time steering}
\label{sec:steering}
At inference we add a scalar multiple of $v_\ell$ to the output of decoder layer $\ell$ at every position,
\begin{equation}
h'_\ell \;=\; h_\ell \;+\; \alpha\, v_\ell ,
\end{equation}
where $\alpha$ is the steering coefficient. Negative values suppress tool use, $\alpha=0$ recovers the unmodified model, and positive values encourage tool use. We select the layer and operating range on a held-out set with the proxy of Eq.~\eqref{eq:proxy} and confirm with real generation.

\paragraph{Projection interventions.}
Beyond additive steering, we apply two projection interventions from the steering literature to test whether $v$ carries the decision rather than merely moving along it. Clamping sets the projection of every position's residual onto $\hat v$ to a fixed target $c$, and directional ablation removes the component entirely ($h'=h-(h\cdot\hat v)\hat v$, the special case $c{=}0$). If the projection onto $\hat v$ is the dial for tool-use propensity, clamping should move the call rate monotonically and ablation should land at a predictable point on the same dial. We verify both in \S\ref{sec:vecanalysis}.

\paragraph{Testing tool-generality.}
Because $v_\ell$ is extracted from a set of several tools, we can ask whether it generalizes to tools that were not in that set. We apply it unchanged to tools absent from the harness (translation, weather, unit conversion, an e-mail action, SQL, and stock lookup) and compare its effect to a vector extracted for each tool in isolation; when comparing vectors of different norm we rescale each to a common norm before applying $\alpha$, so that steering strengths are comparable. A shared whether-to-call direction should steer the held-out tools at or above the strength of their own vectors (\S\ref{sec:vecanalysis}).



%% file: text/experiment.tex
\section{Experiments}
\label{sec:experiment}

We evaluate whether representation steering can reliably control tool use in both under-use and over-use settings. Our experiments examine its effects on tool-call rates, knowledge selectivity, the trade-off between cost and accuracy, and generalization across models. Unless otherwise noted, we use the whether-to-call direction extracted from the multi-tool harness.

\begin{figure*}[t]
\centering
\includegraphics[width=\textwidth]{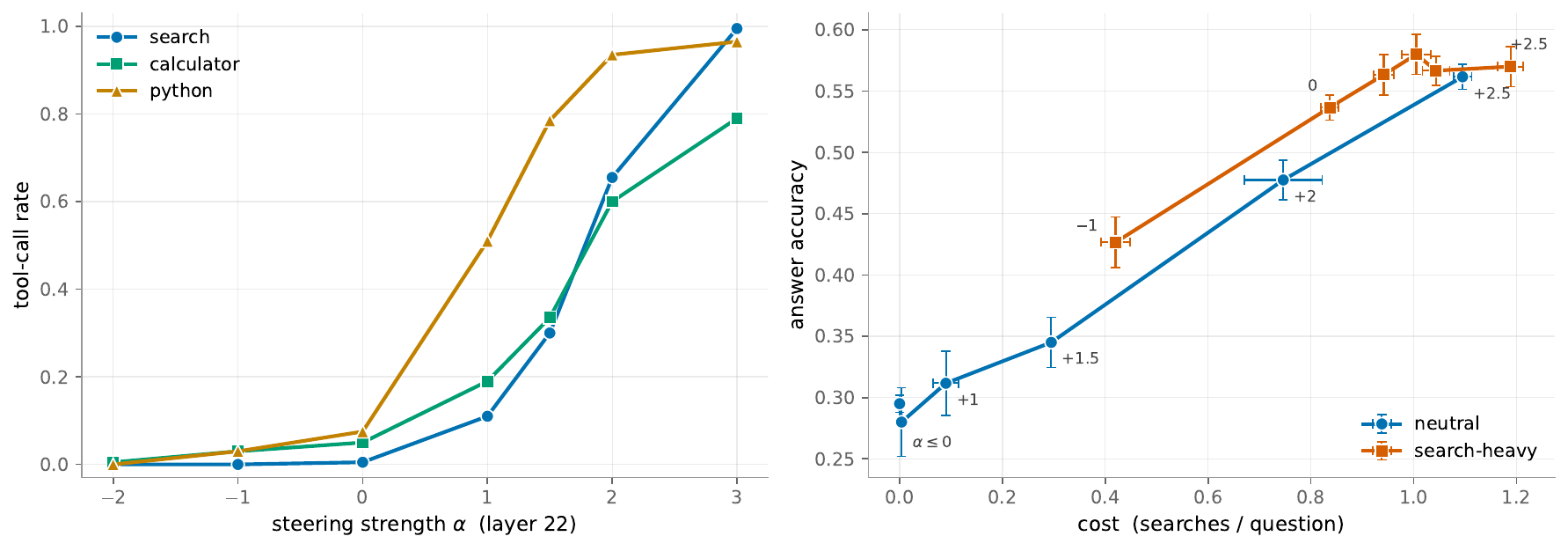}
\caption{A single steering direction provides tool-general control and improves the cost--accuracy trade-off on Qwen3-4B. Left figure shows a single direction, extracted once in the multi-tool harness, reliably controls the tool-call rate when deployed on search, a calculator, or Python. Right figure shows that on PopQA dataset, sweeping $\alpha$ traces a cost/accuracy Pareto frontier.}
\label{fig:headline}
\end{figure*}

\subsection{Experimental Setup}
\label{sec:setup}

\paragraph{Datasets}
\label{sec:datasets}
We evaluate on three public datasets. PopQA~\cite{mallen-etal-2023-trust} contains open-domain factual questions about long-tail entities, which typically require external knowledge to answer correctly. GSM8K~\cite{cobbe2021gsm8k} contains grade-school arithmetic word problems that require precise computation. For string and date manipulation we use two BIG-Bench Hard suites~\citep{suzgun2022challenging}, word sorting and date understanding, single-round tasks with exact answers that reward code execution.
The three datasets probe complementary needs, so a whether-to-call intervention must add calls where they help without inflating calls on questions the model already answers well. PopQA additionally provides a per-entity Wikipedia popularity score, which we use as a proxy for how likely the corresponding fact is to appear in a model's parametric knowledge. To extract steering vectors, we use a few thousand queries drawn in equal parts from PopQA, GSM8K, and the BBH suites. 

\paragraph{Held-out tools and query generation.}
For the tool-generalization study, held-out tools such as translation, weather, unit conversion, e-mail, SQL, and stock lookup receive $300$ tool-matching queries, $200$ used to extract the tool's own vector and $100$ held out for evaluation, plus $200$ non-matching queries drawn from the main pool. The matching queries are generated from LLM-generated templates, short natural requests instantiated from per-tool templates with random fillers such as cities, languages, phrases, unit pairs, and ticker symbols. These queries do not need to be natural or verifiable but instead unambiguous about which tool they call. The experiments measure the propensity to emit a tool call rather than answer accuracy, so a query only needs to make the intended call unambiguous, as in a weather question that a competent agent should route to the weather tool, while the non-matching queries supply the contrast that should not trigger a call. 

\paragraph{Evaluated agents.}
\label{sec:agents}
We evaluate five models spanning dense, MoE, and multimodal architectures: Qwen3-4B-Instruct-2507, Qwen3-8B, Qwen3-30B-A3B-Instruct-2507 (MoE)~\citep{yang2025qwen3technicalreport}, Gemma-4-E4B-it (multimodal)~\citep{gemmateam2026gemma4technicalreport}, and gpt-oss-20b~\citep{openai2025gptoss120bgptoss20bmodel}. 
An evaluated agent combines one of these models with our tool harness. The harness provides the model with a web \emph{search} tool, a \emph{calculator}, and a \emph{Python} coding tool, matched to PopQA, GSM8K, and the BBH suites respectively. For the tool-generalization study, the harness is extended with six held-out tools that never appear during extraction: translation, weather, unit conversion, e-mail, SQL, and stock lookup.

\paragraph{Prompts and baselines.}
\label{sec:prompts}
All experiments share one tool-ambivalent neutral system prompt, which permits but does not require tool use. On PopQA we additionally test a search-heavy system prompt that encourages search, probing the over-use regime. Our baseline is the unsteered model ($\alpha{=}0$) under these prompts, against which we also compare prompt engineering. The full prompt templates are given in the Appendix.

\paragraph{Evaluation setup}
\label{sec:protocol}
We evaluate 200 questions per dataset (600 in total) and 100 template queries per held-out tool. Note that the PopQA sample is drawn from the ten percent of questions with the lowest entity popularity.  Each model uses its own steering vector and operating layer, obtained as described in \S\ref{sec:method}. For each steering strength, the agent answers every evaluation question, and a response counts as a tool call when it emits the model's tool-call token. The per-tool sweeps use $\alpha \in \{-2,-1,0,+1,+1.5,+2,+3\}$, the live frontier uses $\alpha \in \{-1,0,+1,+1.5,+2,+2.5\}$ under both system prompts, and the cross-model comparison uses $\alpha \in \{-2,-1,0,+1,+2\}$.

To characterize the cost/accuracy trade-off, we evaluate the agents with live tool execution, running search against a web API, while running the calculator and Python tools locally. We report answer accuracy and the average number of tool calls per question, with both metrics presented as mean $\pm$ standard deviation over three random seeds.

\begin{figure*}[t]
\centering
\includegraphics[width=\textwidth]{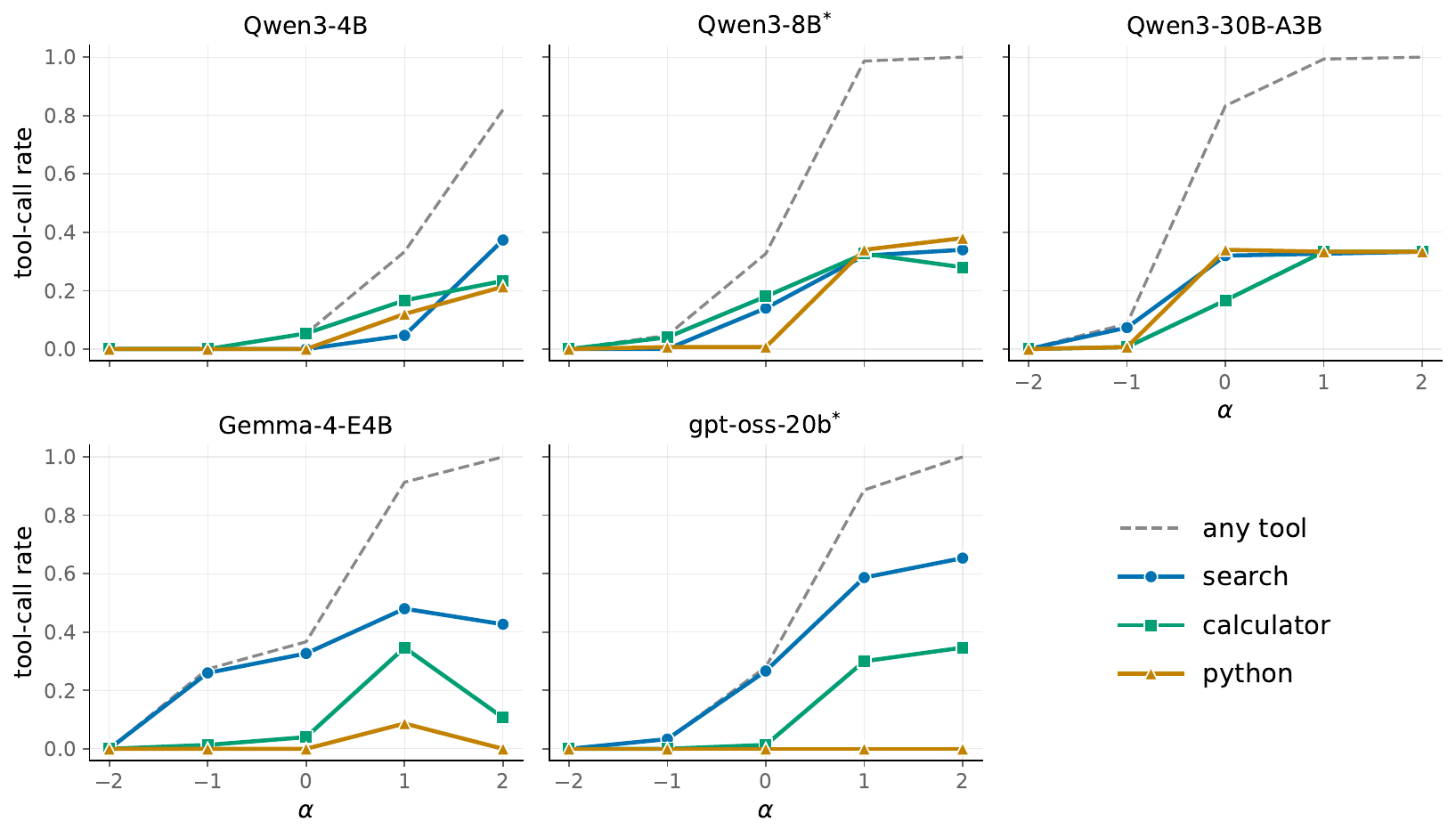}
\caption{Cross-model reproducibility of whether-to-call steering. Across five models covering dense, mixture-of-experts (MoE), and multimodal architectures, increasing the steering coefficient $\alpha$ consistently raises the tool-call rate from near zero to near one. The distribution of calls among search, calculator, and Python varies across models. An asterisk indicates that reasoning was bypassed.}
\label{fig:models}
\end{figure*}

\begin{figure}[t]
\centering
\includegraphics[width=\columnwidth]{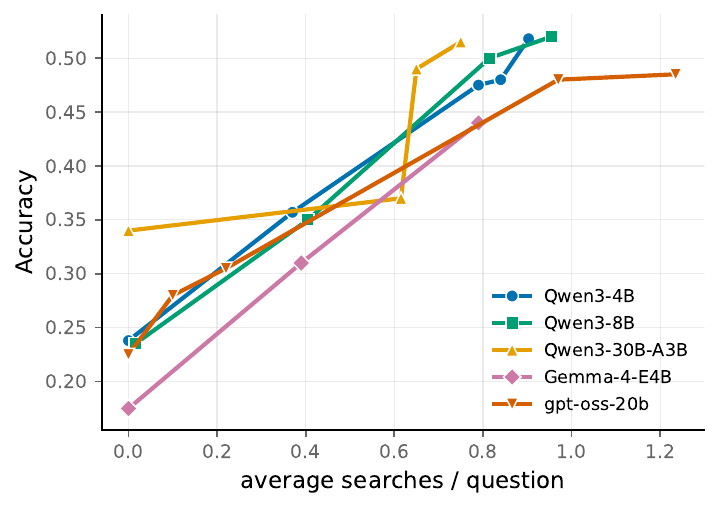}
\caption{Cost-accuracy Pareto frontiers under live search across five models from
three model families. Each point corresponds to one combination of system
prompt and steering coefficient $\alpha$ on PopQA. For clarity, we show only
the non-dominated configurations for each model, so the number of plotted points
differs across models.}
\label{fig:scaling}
\end{figure}

\subsection{Main results}
\paragraph{The direction controls the tool-call rate.}
We first measure how steering changes the tool-call rate, running sampled generation in each of the three single-tool environments at every steering strength. As $\alpha$ increases from $-2$ to $+3$, the call rate rises monotonically from near zero to $0.79$--$1.0$ in all three environments, and the emitted calls remain well-formed throughout this range, as shown in Fig.~\ref{fig:headline}. At $\alpha{=}{+}3$, however, the environment with the highest baseline tool-use tendency begins to produce malformed calls, which marks the edge of the operating window. A single vector, extracted once in the multi-tool harness, therefore controls the call rate of all three tool environments.

\begin{figure*}[t]
\centering
\includegraphics[width=\textwidth]{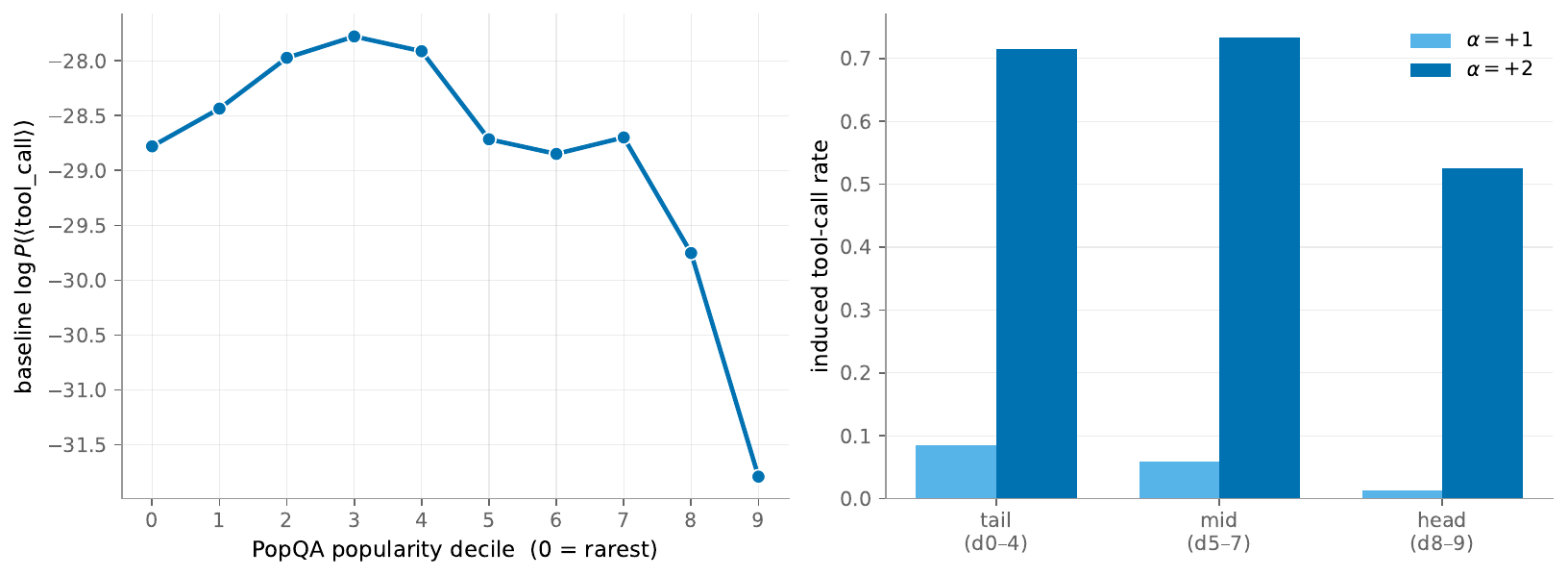}
\caption{Tool-call behavior on PopQA for Qwen3-4B. Left: baseline log-probability of initiating a tool call across entity-popularity deciles. Right: tool-call rates under positive steering
for different entity-popularity questions. }
\label{fig:knowledge}
\end{figure*}

\paragraph{The induced calls target the right questions.}
We next ask where the induced calls land, using the PopQA popularity label as a proxy for how likely the corresponding fact is to appear in a model's parametric knowledge. As shown in Fig.~\ref{fig:headline} (right), the model answers only $29\%$ of questions correctly  without steering, yet it calls the search tool on nearly none of them. Furthermore, as shown in Fig.~\ref{fig:knowledge}, its first-token call tendency is nearly flat across popularity deciles. With positive steering, the new calls concentrate on the low popularity questions that the model cannot answer rather than on the popular head it already knows. The direction therefore recruits tool use exactly where it is needed.

\paragraph{Steering improves the cost--accuracy trade-off.}
Finally, we evaluate whether steering improves answer accuracy when the agent can execute live search. On PopQA, steering the multi-tool agent traces a smooth Pareto frontier between cost and accuracy, as shown in Fig.~\ref{fig:headline}. Accuracy rises from $0.29$ with no searches to $0.56$ at about $1.1$ searches per question. Steering also composes with prompt engineering, and combining it with the search-heavy prompt reaches $0.58$.

\subsection{Generalization across models}

\paragraph{Tool-use steering generalizes across model families and architectures.}
We further show that the tool-use propensity direction can be isolated and steered in models of varying families, sizes, and architectures, applying the same extraction and steering procedure to the other four agents. As shown in Fig.~\ref{fig:models}, baseline call rates span both tool-underuse and tool-overuse, from $0.07$ on Qwen3-4B to $0.83$ on the 30B MoE. Despite this spread, $\alpha{=}{-}2$ suppresses every model to $0.00$, and $\alpha{=}{+}2$ raises four of the five models to $1.00$ and Qwen3-4B to $0.82$. The 30B MoE, an over-user by default, moves in both directions. 

\paragraph{Steering changes whether to call while preserving model-specific routing.}
The per-tool curves in Fig.~\ref{fig:models} show how the additional calls induced by steering are distributed across tools. For the three Qwen models, the call rates of search, calculator, and Python increase together with $\alpha$. At $\alpha{=}{+}2$, the Qwen models approach an approximately balanced routing distribution, with the 30B MoE assigning about one third of the calls to each tool, consistent with the balanced task mixture. The routing pattern is different for Gemma and gpt-oss, where both models rarely select Python at any steering strength and instead route most string and date questions to search or calculator. Thus, steering primarily controls whether the model makes a tool call, while the distribution of calls across tools remains model-dependent. This extends the dissociation in Table~\ref{tab:dissociation} between tool-use propensity and tool choice to other model families.
At large positive $\alpha$, Gemma and gpt-oss also exhibit a gap between the dashed any-tool rate and the sum of the three valid per-tool rates. Inspection shows that these additional calls contain invalid or unrecognized tool names and therefore cannot be assigned to one of the three tools. This is an over-steering failure mode analogous to the malformed calls observed at $\alpha{=}{+}3$ for the primary model.

\paragraph{Steering improves live-search accuracy across models.}
Fig.~\ref{fig:scaling} shows the accuracy and search cost of each model on live-search PopQA, steered with its own multi-tool vector. Every frontier rises monotonically, from a zero-search accuracy of $0.18$--$0.34$ to a peak of $0.44$--$0.52$ at $0.75$--$1.2$ searches per question, a $1.5$--$2.5\times$ gain, and accuracy on searched questions exceeds accuracy on the rest. The frontiers differ mainly in where they start. The 30B MoE has the highest zero-search accuracy $0.34$ and therefore gains the least, while Gemma starts lowest ($0.18$) and remains lowest at its peak $0.44$. Settings that add searches without adding accuracy are dominated and leave the envelope, which is why the models keep different numbers of points.


%% file: text/vecanalysis.tex
\section{Analysis of the Tool-Call Direction}
\label{sec:vecanalysis}

\begin{figure}[t]
\centering
\includegraphics[width=\columnwidth]{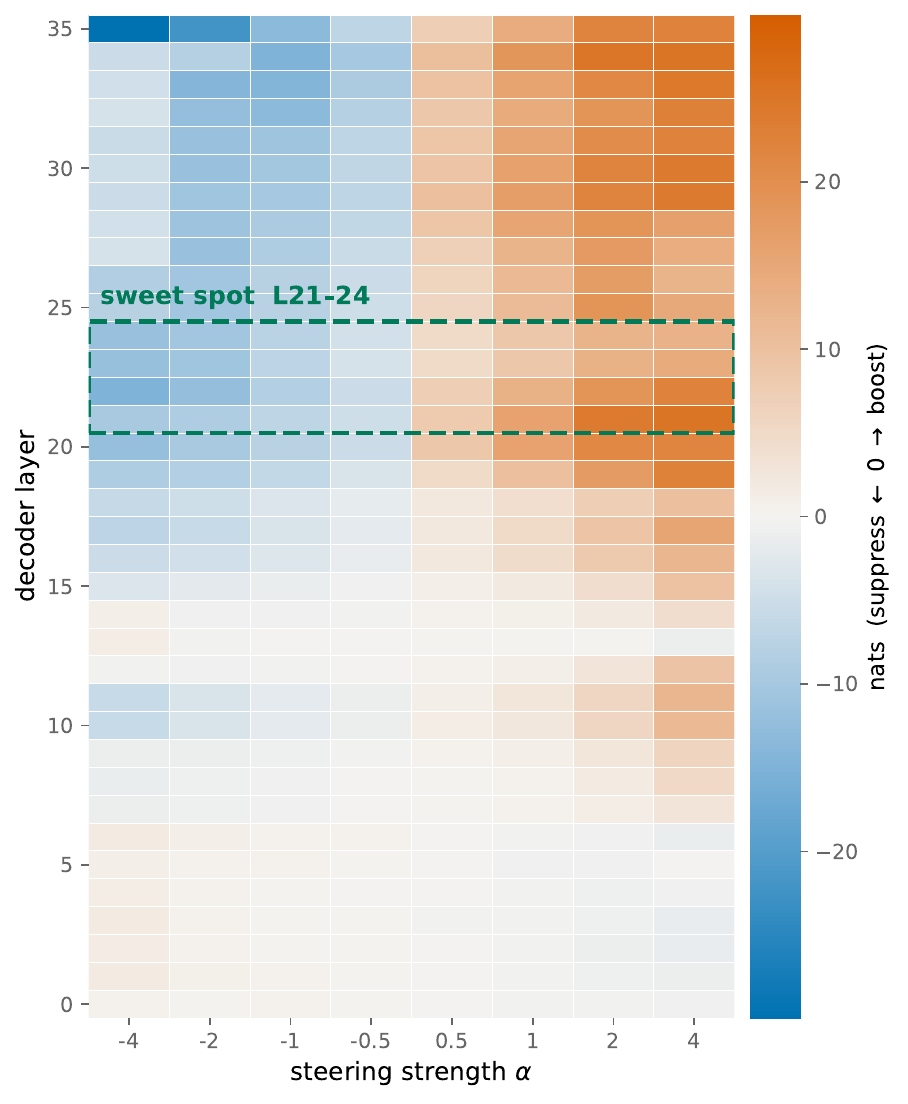}
\caption{Layer $\times$ $\alpha$ steering map ($\Delta\log p(t^{\star})$) on Qwen3-4B. Control is
monotone and strongest in a mid-late band and the last layers are large but erratic.}
\label{fig:layeralpha}
\end{figure}

The previous section showed that the extracted direction controls tool use across tasks
and models. We now analyze the direction itself and answer the following four questions. First, is the direction specific to the tools it was extracted with? We steer six held-out tools and
compare against each tool's own extracted direction in Table~\ref{tab:heldout}. Second,
does it change which tool the model selects? We track per-dataset routing under steering, reported in Table~\ref{tab:dissociation}. Third, where in the network does the direction act, and
over what range of strengths? We map the steering effect at every decoder layer in Fig.~\ref{fig:layeralpha}. Fourth, does the projection onto the direction carry the decision itself? We test the clamping and ablation interventions of \S\ref{sec:steering} in Table~\ref{tab:projection}.

\begin{table}[t]
\centering
\caption{Suppression strength on six held-out tools. Each entry is the drop in the log-probability of the tool-call token when steering with $\alpha{=}{-}2$ in that tool's single-tool harness, $|\Delta\log P|$ in nats averaged over its 100 template queries, so larger means stronger control. $v_{\mathrm{multi}}$ is the direction extracted in the three-tool harness, which never saw these tools, and $v_{\mathrm{own}}$ is the direction extracted from the tool's own queries, rescaled to the same norm.}
\label{tab:heldout}
\begin{tabular}{lcc}
\toprule
Held-out tool & $v_{\mathrm{multi}}$ & $v_{\mathrm{own}}$ \\
\midrule
translate        & 25.3 & 18.4 \\
weather          & 20.1 & 18.8 \\
unit converter   & 26.8 & 23.1 \\
e-mail           & 22.4 & 16.7 \\
SQL              & 20.2 & 22.9 \\
stock price      & 23.9 & 16.6 \\
\bottomrule
\end{tabular}
\end{table}

\paragraph{Tool-general, not tool-specific}
We evaluate whether the tool-use direction extracted from the multi-tool harness transfers to tools that are absent from the extraction data and introduced in out-of-distribution harness configurations. With all vectors norm-matched and the intervention coefficient fixed at $\alpha=-2$, the multi-tool direction substantially reduces the log-probability of the tool-call token for all six held-out tools: translation, weather, unit conversion, e-mail, SQL, and stock-price lookup. It is stronger than each tool's separately extracted direction for five of the six tools and is within $12\%$ of the tool-specific direction for SQL. 
The strong held-out transfer provides evidence that these directions share a domain-general component associated with whether to call a tool, alongside components specific to individual tools or harnesses.

\begin{table}[t]
\centering
\caption{Increasing the steering strength raises the overall tool-call rate while largely preserving task-appropriate tool selection. A dash marks cells with no tool calls.}
\label{tab:dissociation}
\begin{tabular}{lcc}
\toprule
 & baseline & steered ($\alpha{=}{+}2$) \\
\midrule
Total call rate & 0.07 & 0.82 \\
PopQA calls $\to$ search           & --- & 100\% \\
GSM8K calls $\to$ calculator       & 100\% & 99\%  \\
code calls $\to$ Python            & --- & 90\%  \\
\bottomrule
\end{tabular}
\end{table}

\paragraph{Operating layer and steering strength}
As shown in Fig.~\ref{fig:layeralpha}, we map where in the network the direction is effective by adding $\alpha v_\ell$ at each decoder layer $\ell$ in turn and measuring the resulting change in $\log p(t^{\star})$ on held-out prompts from the three-tool harness. We sweep $\alpha\in[-4,4]$ on Qwen3-4B. Early layers are inert, and below $L7$ the intervention moves the propensity by less than $2$ nats even at $\alpha{=}{\pm}4$. Control grows through the middle of the stack and is strongest, and monotone in $\alpha$, in a mid-late band ($L21$--$24$). At the operating layer $L22$ the propensity spans $37$ nats across $\alpha\in[-4,4]$, roughly sixteen orders of magnitude, and about $80\%$ of this range is already reached at $\alpha{=}{\pm}2$. Above the band the effect remains large but is no longer monotone. At \ $L27$, for example, $\alpha{=}{-}2$ suppresses the
propensity by $11.6$ nats while $\alpha{=}{-}4$ suppresses it by only $3.9$, so doubling
the strength partially undoes the effect, and the final layers show the largest raw swings
but the least reliable ones. This pattern is consistent with the whether-to-call decision
being computed at mid-late depth. Before the band the representation the direction targets
has not yet formed, and past the band the intervention increasingly acts on the output
head rather than on the decision.

\begin{table}[t]
\centering
\caption{Tool-call rates under projection clamping and directional ablation for an under-user, Qwen3-4B $L22$, and an over-user, Gemma-4-E4B $L35$, evaluated on 150 mixed questions. Clamp targets are standardized relative to each model's baseline projection distribution, with $0\sigma$ denoting its mean. The no-intervention row reports the model's natural call rate. Directional ablation sets the raw projection onto $\hat v$ to zero, corresponding to $+2.1\sigma$ for Qwen3-4B and $+1.3\sigma$ for Gemma-4-E4B.}
\label{tab:projection}
\begin{tabular}{lcc}
\toprule
Clamp target $c$ & Qwen3-4B & Gemma-4-E4B \\
\midrule
no intervention & 0.05 & 0.37 \\
\addlinespace
$-2\sigma$ & 0.01 & 0.01 \\
$-1\sigma$ & 0.02 & 0.17 \\
$0\sigma$ (baseline mean) & 0.03 & 0.61 \\
$+1\sigma$ & 0.06 & 0.87 \\
$+2\sigma$ & 0.11 & 1.00 \\
$+3\sigma$ & 0.16 & 1.00 \\
\addlinespace
ablation (raw $c{=}0$) & 0.11 & 0.90 \\
\bottomrule
\end{tabular}
\end{table}

\paragraph{Projection interventions}
We evaluate the clamping and ablation interventions introduced in \S\ref{sec:steering} on an under-user (Qwen3-4B) and an over-user (Gemma-4-E4B), using the selected direction and intervention layer for each model. As shown in  Table~\ref{tab:projection}, increasing the clamped projection monotonically increases tool use in both models. The effect is strong for Gemma-4-E4B, whose call rate rises from $0.01$ to $1.00$, but substantially weaker for Qwen3-4B, whose call rate rises only from $0.01$ to $0.16$. Thus, the projection causally affects the whether-to-call decision in both models, but clamping does not recover the full control range obtained with additive steering: on Qwen3-4B, additive steering raises the call rate to $0.82$.

One likely explanation is that the two interventions preserve different information. Additive steering shifts each query's activation while retaining query-dependent variation along $\hat v$. Clamping instead assigns the same projection value to every query, discarding this variation. If tool-call probability depends jointly on the projection and query-specific activation structure, fixing the projection alone may produce only a limited change in an under-user such as Qwen3-4B. This also explains why clamping to the mean of the baseline projection distribution does not necessarily reproduce the no-intervention call rate. The difference is small for Qwen3-4B but substantial for Gemma-4-E4B: without intervention, projections vary across queries, whereas the $0\sigma$ clamp replaces the entire baseline distribution with a single value.

Directional ablation removes the component along $\hat v$ and therefore sets the raw projection to zero. Because the mean baseline projection is negative in both models, raw zero corresponds to $+2.1\sigma$ for Qwen3-4B and $+1.3\sigma$ for Gemma-4-E4B. The resulting ablation call rates, $0.11$ and $0.90$, are consistent with the corresponding locations on the clamp curves.

%% file: text/rw.tex
\section{Related Work}
\label{sec:rw}

\paragraph{Linear representations and geometry.}
The linear representation hypothesis~\citep{park2024linearrepresentationhypothesisgeometry}
predicts that concepts encoded in model weights should be linearly decodable from
representations, with supporting evidence from word-vector arithmetic~\citep{mikolov2013linguistic}
and superposition theory~\citep{elhage2022toymodelssuperposition}. Empirical work on
refusal~\citep{arditi2024refusallanguagemodelsmediated} and
truthfulness~\citep{burns2024discoveringlatentknowledgelanguage} has found steering vectors
with clean geometric properties consistent with this account, while more recent work
complicates the picture: concept directions can span multi-dimensional subspaces rather than
single directions~\citep{wollschlager2025geometryrefusallargelanguage,
siu2025repitrepresentingisolatedtargets}, with geometry that depends on context and layer
depth. Our setting differs from these studies in one respect. The concepts they steer
are encoded in the model's weights, while tools are injected through context,
so it is not obvious a priori that a context-dependent behavior should admit a
single weight-space steering direction. Our results show that the
whether-to-call decision nonetheless admits one. A vector extracted in one
harness steers six held-out tools at or near the strength of each tool's own
vector (Table~\ref{tab:heldout}), consistent with a shared whether-to-call
component alongside tool- and harness-specific structure.

\paragraph{Representation steering.}
Steering methods identify directions in representation space corresponding to target behaviors and
modulate them via vector addition or orthogonalization~\citep{zou2023representationengineeringtopdownapproach,
arditi2024refusallanguagemodelsmediated, turner2024steeringlanguagemodelsactivation,
panickssery2024steeringllama2contrastive, siu2025steeringsafetysystematicsafetyevaluation}.
Directions are commonly extracted from contrastive data
pairs~\citep{burns2024discoveringlatentknowledgelanguage, arditi2024refusallanguagemodelsmediated,
zou2023representationengineeringtopdownapproach} and have been used for both behavior elicitation
and concept removal~\citep{ravfogel2020nulloutguardingprotected, cosmic}; inference-time variants
use probes or classifiers to apply interventions conditionally during the forward
pass~\citep{li2023inference, lee2025programmingrefusalconditionalactivation}. Prior steering work
has largely assumed that the concept being steered has stable parametric grounding. We instead
apply the same lightweight difference-of-means machinery to a discrete, context-dependent agentic
decision, whether to emit a tool call, and turn it into a continuous, monotone inference-time
knob whose effect we validate not only on a forward-pass proxy but end-to-end on task
cost and accuracy.

\paragraph{LLM tool use and deciding when to call.}
Tool-use methods extend language models with external functions, APIs, and environments, granting
access to up-to-date information, specialized computation, and domain expertise beyond parametric
memory~\citep{schick2023toolformer, yao2023reactsynergizingreasoningacting,
qin2023toolllmfacilitatinglargelanguage, Qu_2025, shen2024llmtoolssurvey}. A body of work studies
how models decide \emph{when} to call tools, \emph{which} tools to select, and \emph{how} to
incorporate tool outputs into subsequent reasoning~\citep{qian2024toolinklinkingtoolkitcreation},
and improves robustness through tool creation, module integration, and alignment for more efficient
calling under uncertainty or knowledge-boundary awareness~\citep{qian2024creatortoolcreationdisentangling,
liu2025toolacewinningpointsllm, qian2024investigateconsolidateexploitgeneralstrategyintertask,
xu-etal-2025-alignment}. Dedicated evaluations show that models still make unnecessary or incorrect
tool calls, motivating benchmarks for deciding whether a tool is needed and which should be
used~\citep{huang2024metatoolbenchmarklargelanguage, ning2024wtuevalwhetherornottoolusage}, and a
closely related line targets tool overuse and adaptive calling based on model uncertainty or
self-awareness~\citep{wang-etal-2025-self, shen-etal-2024-smartcal, qian-etal-2025-smart}. These
methods largely optimize an external tool-interaction policy through training or prompting; in
contrast, we induce the desired whether-to-call behavior through a training-free, inference-time
manipulation of internal representations, covering both the under-use and over-use regimes with a
single continuous knob and, unlike prior over-use mitigations, without task-specific supervision.

%% file: text/conclusion.tex
\section{Conclusion}
\label{sec:conclusion}
We showed that whether an LLM agent calls a tool can be controlled by a single linear direction in the residual stream. The direction is extracted using forward passes alone, and adding it at one mid-to-late layer moves the call rate continuously from near zero to near one, without training or prompt changes. Three properties make this control useful beyond simply setting the call rate. It is knowledge-selective, concentrating induced calls on questions the model cannot answer on its own. It is tool-general, steering six held-out tools at or near the strength of their own separately extracted directions. It also changes whether the model calls without disturbing which tool it selects. Under live tool execution, sweeping the steering coefficient traces a cost--accuracy Pareto frontier that nearly doubles open-domain QA accuracy. The same procedure transfers to five models from three vendors whose baselines range from under-use to over-use.

The method also has clear limitations. It controls the decision to call a tool, but not the quality of tool execution. For reasoning-first models, the reasoning span must be bypassed before the decision becomes readable. In addition,  strong positive steering can corrupt the tool-call format. Despite these limitations, the tool-use direction provides a lightweight inference-time mechanism for controlling the cost and reliability of tool-using agents.

%% file: appendix.tex
\section{Prompt Templates}
\label{sec:supp-prompts}

All experiments use one of the two system prompts below, identical strings
for all models. The \emph{neutral} prompt, used everywhere unless stated
otherwise, permits but does not require tool use:

\begin{quote}\small\ttfamily
You are a helpful assistant. You may use the provided tools when they are
helpful for answering the user's question, but you are not required to use
them.
\end{quote}

The \emph{search-heavy} prompt supplies the over-use pole for the
cost--accuracy frontier on PopQA:

\begin{quote}\small\ttfamily
You are a helpful assistant with access to a web search tool. Many factual
questions concern long-tail knowledge that is easy to get wrong from memory,
so you are encouraged to use the search tool to check facts before answering.
\end{quote}

User turns contain the evaluation question verbatim, with no additional
instructions or few-shot examples.

\section{Tool Harness Details}
\label{sec:supp-harness}

\paragraph{Tool-call opener.}
The propensity readout uses the tool-exclusive token that opens an assistant
tool call, shown in Table~\ref{tab:supp-tstar}. For the harmony-format model,
whose tool route is a channel choice rather than a single opener token, the
propensity is read contrastively at the decision position, comparing the
tool-call route against the direct-answer route.

\begin{table}[t]
\centering
\caption{Tool-call opener $t^{\star}$ per model family.}
\label{tab:supp-tstar}
\begin{tabular}{ll}
\toprule
Model family & Opener \\
\midrule
Qwen3 & \texttt{<tool\_call>} \\
Gemma-4 & \texttt{<|tool\_call>} \\
gpt-oss & channel-route readout (no single opener) \\
\bottomrule
\end{tabular}
\end{table}

\paragraph{Tool schemas.}
Tables~\ref{tab:supp-harness-tools} and~\ref{tab:supp-heldout-tools} list the
function schemas passed to the chat template. Descriptions are given to the
model verbatim. All schemas follow the standard JSON function-calling format
with the listed arguments marked required.

\begin{table}[t]
\centering
\footnotesize
\caption{Tools of the multi-tool harness.}
\label{tab:supp-harness-tools}
\begin{tabular}{p{0.23\columnwidth}p{0.23\columnwidth}p{0.36\columnwidth}}
\toprule
Tool & Arguments & Description \\
\midrule
\texttt{search} & \texttt{query} & Search the web and return the top results. Useful for up-to-date or long-tail factual information. \\
\texttt{calculator} & \texttt{expression} & Evaluate an arithmetic expression and return the numeric result. \\
\texttt{python} & \texttt{code} & Execute a short Python snippet and return its stdout. \\
\bottomrule
\end{tabular}
\end{table}

\begin{table}[t]
\centering
\footnotesize
\caption{Held-out tools, listed under the function identifiers the model
sees. The main paper refers to the same six tools, in this order, by task
name: translation, weather, unit conversion, e-mail, SQL, and stock lookup.}
\label{tab:supp-heldout-tools}
\begin{tabular}{p{0.26\columnwidth}p{0.26\columnwidth}p{0.3\columnwidth}}
\toprule
Tool & Arguments & Description \\
\midrule
\texttt{translate} & \texttt{text},\newline \texttt{target\_\allowbreak language} & Translate text between languages. \\
\addlinespace
\texttt{get\_weather} & \texttt{location}, \texttt{when} & Get the current or forecast weather for a location. \\
\addlinespace
\texttt{unit\_\allowbreak converter} & \texttt{value}, \texttt{from\_unit}, \texttt{to\_unit} & Convert a quantity between units of measurement. \\
\addlinespace
\texttt{send\_email} & \texttt{to}, \texttt{subject}, \texttt{body} & Send an email on the user's behalf. This action is irreversible. \\
\addlinespace
\texttt{run\_sql} & \texttt{query} & Run a read-only SQL query against the company database (tables: customers, orders, products). \\
\addlinespace
\texttt{stock\_price} & \texttt{ticker} & Look up the latest stock price for a ticker symbol. \\
\bottomrule
\end{tabular}
\end{table}

\section{Sampling and Query Construction}
\label{sec:supp-data}

\paragraph{Extraction pool.}
The extraction pool is uniform over PopQA entity popularities and disjoint
from all evaluation sets.

\paragraph{Popularity-decile study.}
For the knowledge-selectivity analysis of the main paper, we sort PopQA by
entity popularity, draw $40$ questions per popularity decile, and render each
in the single-tool search environment under the neutral prompt. The
first-token propensity panel reports the mean $\log p(t^{\star})$ per decile
with no generation. The call-rate panel reports realized tool-call rates from
sampled generation, grouped into tail (deciles 0--4), mid (5--7), and head
(8--9). The tail/head comparison of the projection study uses $100$ rare-tail
and $100$ head questions in the same environment. The popularity score counts
Wikipedia page views of the question's subject entity. Head questions ask
about heavily viewed subjects, as in ``What is the capital of India?''
($s_{\mathrm{pop}} \approx 1.3 \times 10^{6}$), while tail questions ask about
sparsely viewed ones, as in ``In what country is Valea Seac\u{a} River?''
($s_{\mathrm{pop}} = 2$) or ``What sport does Miroslav Milutinovi\'c play?''
($s_{\mathrm{pop}} = 28$).

\section{Hyperparameters and Protocols}
\label{sec:supp-hparams}

\paragraph{Contrastive pools.}
The top and bottom quantiles of the main paper are the top and bottom $10\%$
by $s(q)$ within each question type, so both pools are balanced across
question types.

\paragraph{Generation.}
Call rates are measured with sampled generation at temperature $0.7$,
top-$p$ $0.8$, and at most $384$ new tokens, with a fixed seed for the
call-rate sweeps.

\paragraph{Failure criteria.}
The called tool is identified by parsing the tool name from the call payload,
and a call whose payload names no recognized tool is counted as malformed. A
response is counted as degenerate when it is empty or repeats a short window
of text at least five times near its start, the criterion behind the reported
edge of the operating window.

\section{Projection-Intervention Curves}
\label{sec:supp-projection}

\begin{figure*}[t]
\centering
\includegraphics[width=\textwidth]{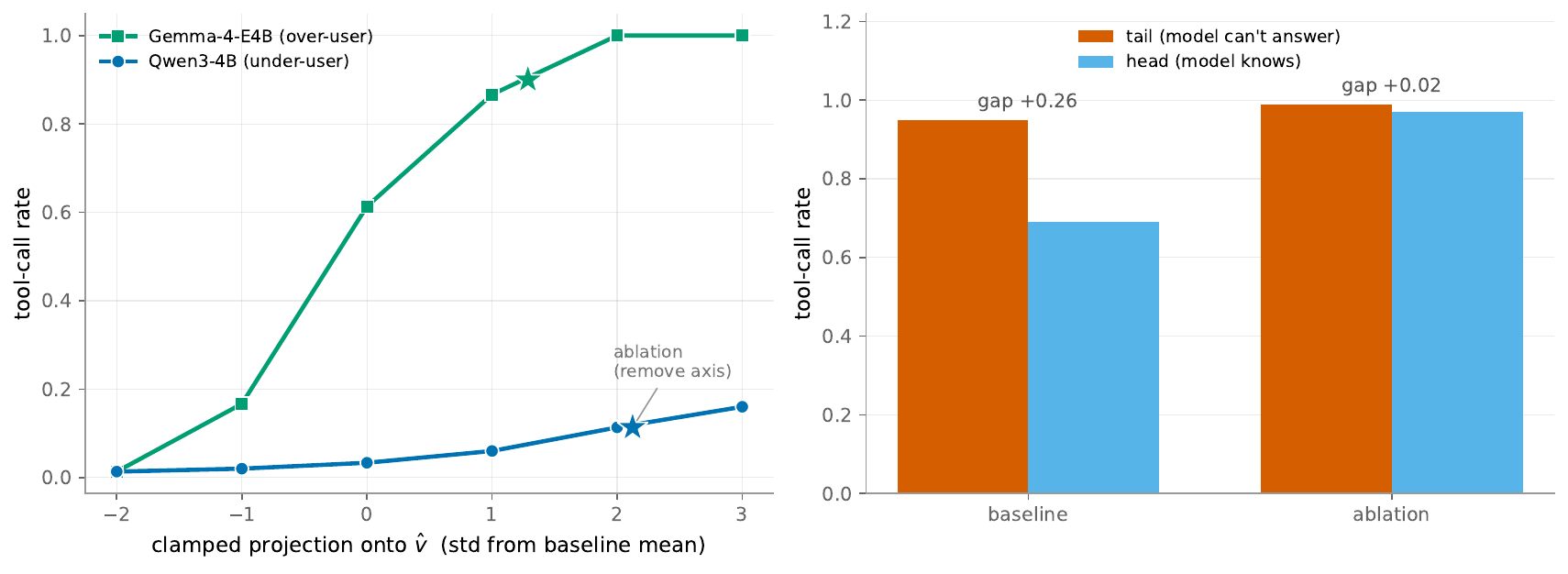}
\caption{Projection interventions. \emph{Left:} clamping the projection onto
$\hat v$ moves the call rate monotonically for both an under-user (Qwen3-4B)
and an over-user (Gemma-4-E4B), and the ablation point ($c{=}0$, stars) lies
on each model's clamp curve. Because the baseline mean projection is negative,
raw zero sits above the baseline operating point on the axis. \emph{Right:}
ablation erases knowledge selectivity, collapsing the tail/head call-rate gap
from $+0.26$ to $+0.02$.}
\label{fig:supp-projection}
\end{figure*}

Fig.~\ref{fig:supp-projection} (left) plots the full clamp curves summarized
by the projection table of the main paper. Clamping sets
$h' = h + (c - h\!\cdot\!\hat v)\,\hat v$ at every position with target
$c = \mu + t\sigma$, where $\mu, \sigma$ are the mean and standard deviation
of baseline last-prompt-token projections onto $\hat v$ over the evaluation
set; directional ablation ($c = 0$) sits at $t = -\mu/\sigma$ on the same
axis. The curves fill in the shape between the grid points of the table. The
over-user's response rises steeply across the baseline range and saturates
above $+2\sigma$, the under-user's remains shallow over the same interval, and
each model's ablation star point  falls on its own clamp curve.

The right panel reports an experiment not included in the main paper,
knowledge selectivity under ablation. On the over-user in the single-tool
search environment, we measure call rates separately on the $100$ rare-tail
and $100$ head questions of the popularity-decile study
(\S\ref{sec:supp-data}). At baseline the model calls search on $0.95$ of tail
questions but only $0.69$ of head questions, a $+0.26$ gap aligned with what
the model knows. After ablation both groups rise to near ceiling ($0.99$ and
$0.97$) and the gap collapses to $+0.02$. Removing the component along
$\hat v$ therefore removes the knowledge-dependent modulation of the call
decision and leaves a nearly indiscriminate caller, so the signal that
distinguishes known from unknown questions travels along this direction.

